\pdfoutput=1

\documentclass[11pt]{article}

\usepackage[final]{acl}

\usepackage{times}
\usepackage{latexsym}

\usepackage[T1]{fontenc}

\usepackage[utf8]{inputenc}

\usepackage{microtype}

\usepackage{inconsolata}

\usepackage{graphicx}
\usepackage{colortbl}
\usepackage{makecell}
\usepackage{amssymb}
\usepackage{amsmath}
\usepackage{lingmacros}
\usepackage{tikz-dependency}
\usepackage{forest}
\usepackage{subcaption}
\usepackage{float}
\usepackage{fontawesome}
\DeclareMathOperator{\edge}{CountEdge}
\DeclareMathOperator{\rmstart}{START}
\DeclareMathOperator{\rmend}{END}
\DeclareMathOperator{\kl}{KL}
\DeclareMathOperator{\tree}{TREE}
\DeclareMathOperator{\nwp}{NWP}
\newlength{\astspace}
\newlength{\excspace}
\newcommand{\un}{*}
\newcommand{\exc}{!}
\newcommand{\ac}{\hspace*{\astspace}}
\newcommand{\acexc}{\hspace*{\excspace}}

\title{Tree-Planted Transformers:\\ Unidirectional Transformer Language Models \\with Implicit Syntactic Supervision}

\author{Ryo Yoshida \and Taiga Someya \and Yohei Oseki \\
  The University of Tokyo \\
  \texttt{\{yoshiryo0617, taiga98-0809, oseki\}@g.ecc.u-tokyo.ac.jp}}

\begin{document}
\maketitle
\begin{abstract}
Syntactic Language Models (SLMs) can be trained efficiently to reach relatively high performance; however, they have trouble with inference efficiency due to the explicit generation of syntactic structures. In this paper, we propose a new method dubbed \textbf{tree-planting}: instead of explicitly generating syntactic structures, we ``plant'' trees into attention weights of unidirectional Transformer LMs to implicitly reflect syntactic structures of natural language. Specifically, unidirectional Transformer LMs trained with tree-planting will be called \textbf{Tree-Planted Transformers (TPT)}, which inherit the training efficiency from SLMs without changing the inference efficiency of their underlying Transformer LMs. Targeted syntactic evaluations on the SyntaxGym benchmark demonstrated that TPTs, despite the lack of explicit generation of syntactic structures, significantly outperformed not only vanilla Transformer LMs but also various SLMs that generate hundreds of syntactic structures in parallel. This result suggests that TPTs can learn human-like syntactic knowledge as data-efficiently as SLMs while maintaining the modeling space of Transformer LMs unchanged.

\begin{center}
    \faGithub~\url{https://github.com/osekilab/TPT}
\end{center}
\end{abstract}
\section{Introduction}
Recent years have witnessed remarkable success in Large Language Models (LLMs) based on Transformer LMs~\citep{vaswani2017Attention}. However, despite their success, Transformer LMs have some drawback in \textit{training efficiency}---especially when compared with humans. For example, GPT-3~\citep{brown2020Language} is trained on around $2,000\times$ larger data than a 12-year-old human would have experienced~\citep{warstadt2023Findings}, indicating that Transformer LMs lack sufficient inductive bias for language acquisition.
\begin{figure}
    \centering
    \includegraphics[width=7.5cm]{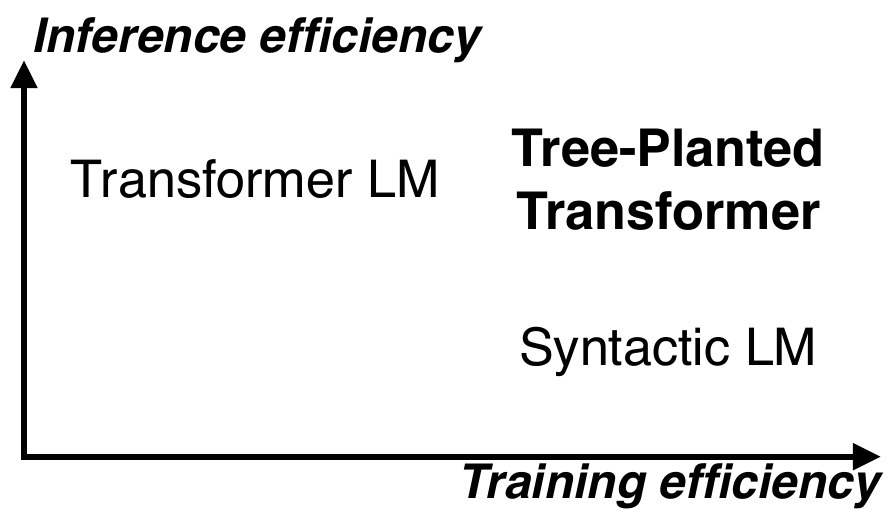}
    \caption{Two types of efficiency: training efficiency and inference efficiency. Our Tree-Planted Transformers (TPT) inherit the training efficiency from Syntactic LMs without changing the inference efficiency of their underlying Transformer LMs.}
    \label{fig:efficiency}
\end{figure}

On another strand, previous work has revealed that Syntactic Language Models (SLMs), defined as a generative models of a token sequence and its syntactic structures, can achieve high syntactic performance under data-constrained settings~\citep{dyer2016Recurrent,noji2021Effective,qian2021Structural,sartran2022Transformer,yoshida2022Composition,murty2023Pushdown}. For example, \citet{sartran2022Transformer} showed that some SLMs can achieve comparable syntactic performance to an LLM-like model\footnote{Due to the rapid advances in recent years, what were once considered LLMs are no longer deemed ``large'' by current standards. We will refer to Transformer LMs larger than or equal to GPT-2~\citep{radford2019Language} as \textit{LLM-like} models.} that is trained with medium---around $250\times$ larger---data, suggesting that syntactic supervision is essential for LMs to achieve high training efficiency. However, despite their training efficiency, SLMs have trouble with \textit{inference efficiency}---they require hundreds of syntactic structures generated via beam search~\citep{stern2017Effective,crabbe2019Variable} or an external parser to precisely approximate marginal distribution over a token sequence, which naturally incurs the costs hundreds of times higher than their underlying sequential models.

In this paper, we propose a new method dubbed \textbf{tree-planting}: instead of explicitly generating syntactic structures, we ``plant'' trees into attention weights of unidirectional Transformer LMs to implicitly reflect syntactic structures of natural language.\footnote{The term ``tree-planting'' coincidentally bears a resemblance to the term used in \citet{mueller-linzen-2023-plant}, but this work diverges from ours in its motivation. Specifically, \citet{mueller-linzen-2023-plant} investigated biases that enable syntactic generalization in Transformer LMs, from the perspectives of architectural features (depth, width, and number of parameters), as well as the genre and size of training corpus.} Specifically, unidirectional Transformer LMs trained with tree-planting will be called \textbf{Tree-Planted Transformers (TPT)}, which inherit the training efficiency from SLMs without changing the inference efficiency of their underlying Transformer LMs (Figure~\ref{fig:efficiency}).

Previous studies have also explored the syntactic supervision of attention weights, mainly targeting bidirectional Transformer Encoders (\citealp{wu2018Phraselevel,DBLP:conf/iclr/NguyenJHS20,bugliarello2020Enhancing,bai2021SyntaxBERT,sachan2021Syntaxa,slobodkin2022Semanticsaware}; \textit{inter alia}) (\S\ref{sec:rel}). These encoder-oriented approaches assume the entire sentence as input and typically aim to reflect the syntactic relationship between input words in a bottom-up manner. In contrast, tree-planting is uniquely designed for unidirectional Transformer LMs that can be used for text generation---and they have recently increased prevalence because of their compatibility with instruct tuning~\citep{zhang2024Instruction} or Reinforcement Learning from Human Feedback (RLHF, \citealp{ouyang2022Training}). Specifically, tree-planting considers syntactic structures involving the next word to be generated, by focusing on the \textit{syntactic distance}~\citep{DBLP:conf/iclr/ShenLHC18,DBLP:conf/iclr/ShenTSC19,du2020Exploiting} between the next word and the previous words in the context (\S\ref{sec:TP}).

Targeted syntactic evaluations on the SyntaxGym benchmark~\citep{gauthier2020SyntaxGym} demonstrated that TPTs, despite the lack of explicit generation of syntactic structures, significantly outperformed not only vanilla Transformer LMs but also various SLMs that generate hundreds of syntactic structures in parallel. This result suggests that TPTs can learn human-like syntactic knowledge as data-efficiently as SLMs while maintaining the modeling space of Transformer LMs unchanged. Furthermore, closer inspection of syntactic phenomena implied that tree-planting shows high compatibility with dependency structures (\S\ref{sec:exp}).

Additionally, we analyzed the two hyperparameters introduced by tree-planting: (i) the number of heads where the tree-planting loss is applied, and (ii) the balance between the next-word prediction loss and the tree-planting loss. Our results demonstrated that the highest accuracy was generally achieved when a single head was adopted as a tree-planted head while the balancing parameter emerged as a crucial hyperparameter; excessively high and low weights on the tree-planting loss both led to ineffective outcomes (\S\ref{sec:analysis}).

\section{Related work}
\label{sec:rel}
\begin{table*}
    \small
    \centering
    \begin{tabular}{lcccc}
        \hline
         &\makecell{Parser-free\\inference}&\makecell{Syntactic \\supervision}&\makecell{Unidirectional\\LM}&\makecell{Parallel\\computation}\\
         \hline
         \makecell[l]{
            \citet{wu2018Phraselevel};\citet{DBLP:conf/iclr/NguyenJHS20};\\
            \citet{bugliarello2020Enhancing};\citet{bai2021SyntaxBERT};\\
            \citet{sachan2021Syntaxa};\citet{slobodkin2022Semanticsaware}
         }& & \checkmark & & \checkmark\\
         \hline
         \citet{wang2019Tree} & \checkmark & & & \checkmark \\
         \makecell[l]{
            \citet{strubell2018LinguisticallyInformed};\citet{chen2023Sudden}
         }& \checkmark & \checkmark & & \checkmark \\
         \citet{peng2019PaLM} & \checkmark & \checkmark & \checkmark & \\
         \hline
         \rowcolor[gray]{0.83} Tree-planting (ours) &\checkmark&\checkmark&\checkmark&\checkmark\\
         \hline
    \end{tabular}
    \caption{Comparison of our tree-planting with the previous work that constrains attention weights according to syntactic structures, based on the requirements for the architecture that inherits the training efficiency of SLMs without changing the inference efficiency of their underlying Transformer LMs: (i) parser-free inference, (ii) syntactic supervision, (iii) unidirectional LM, and (iv) parallel computation.}
    \label{tab:comparison_attn}
\end{table*}

\subsection{Syntactic Language Model}
\label{subsec:slm}
Syntactic Language Models (SLMs) are a generative model of a token sequence $\mathbf{x}$ and its syntactic structure $\mathbf{y}$. Formally, SLMs are defined as:
\begin{equation}
    p(\mathbf{x}, \mathbf{y}) = p(\mathbf{z}) = \prod_{t=1}^{n} p(z_t|z_{<t}) ,
\end{equation}
where $\mathbf{z}$ denotes the sequence of actions to generate both the token sequence and syntactic structure. For example, in top-down and left-to-right SLMs, each $z_t$ could be either generating a token or opening/closing a constituent.

Recently, several SLMs based on the Transformer architecture have been proposed, achieving higher syntactic performance than medium LLM-like models~\cite{qian2021Structural,sartran2022Transformer,murty2023Pushdown}. However, because SLMs generate both a syntactic structure and token sequence, they cannot be directly utilized as LMs, or a generative model of a token sequence. To precisely approximate the marginal distribution over a token sequence, i.e., $p(\mathbf{x}) = \sum_{\mathbf{y} \in \mathcal{Y}} p(\mathbf{x}, \mathbf{y})$ (where $\mathcal{Y}$ represents the set of possible syntactic structures behind $\mathbf{x}$), they require hundreds of syntactic structures generated via beam search~\citep{stern2017Effective,crabbe2019Variable} or an external parser. Although actual costs would depend on the hardware and metrics, the calculation of $p(\mathbf{x}, \mathbf{y})$ for each structure in $\mathcal{Y}$ naturally multiplies inference costs by $|\mathcal{Y}|$ compared to the sequential models that directly calculates $p(\mathbf{x})$. Furthermore, additional costs would be incurred by the beam search procedure or the external parser itself.

\subsection{Constraints on attention weights}
\label{subsec:const}
As discussed in \S\ref{subsec:slm}, the bottleneck that impairs SLMs' inference efficiency is their modeling space of the joint probability. To achieve the architecture that inherits the training efficiency of SLMs without changing the inference efficiency of their underlying Transformer LMs, it is necessary to introduce syntactic supervision without changing the modeling space of the sequential models. For our goal, we will build upon another line of approach that constrains attention weights according to syntactic structures---mainly targeting bidirectional Transformer Encoders like BERT~\citep{bert}.\footnote{Beyond studies that constrain attention weights according to syntactic structures, there are also investigations that aimed at incorporating various types of information, such as word alignment, into the attention mechanisms \citep[e.g.,][]{yin2021Compositional}. This subsection, however, specifically focuses on studies that target syntactic biases.} Table~\ref{tab:comparison_attn} summarizes the previous work in this line of approach, comparing our tree-planting (\S\ref{sec:TP}) against others based on the requirements for our goal: (i) parser-free inference, (ii) syntactic supervision, (iii) unidirectional LM, and (iv) parallel computation.

First, the majority of these approaches are purely motivated to explicitly restrict attention weights with syntactic structures from external parsers, under the assumption that these parsers would be available during inference~\citep{wu2018Phraselevel,DBLP:conf/iclr/NguyenJHS20,bugliarello2020Enhancing,bai2021SyntaxBERT,sachan2021Syntaxa,slobodkin2022Semanticsaware}. These studies achieved successful performance in their respective downstream tasks, but not only are their approaches all not directly applicable to unidirectional LMs, they also require external parsers during inference, rendering them not aligned with our goal of the inference efficient architecture.

Second, several approaches have been proposed that eliminate the need for external parsers during inference, but they still fall short of meeting all the requirements. \citet{wang2019Tree} aimed at an unsupervised approach, where a hierarchical architectural bias widens the range of neighboring tokens eligible to attend from lower to upper layers, yet this method is still not aligned with our goal of achieving higher training efficiency via syntactic supervision. Additionally, \citet{strubell2018LinguisticallyInformed} and \citet{chen2023Sudden} designed the loss functions that implicitly encourage the attention to syntactic parents or children for each token, satisfying the 3/4 requirements for our goal.\footnote{Other than the approach to constrain attention weights, \citet{tziafas-etal-2023-improving} proposed a method to train Transformer Encoders in a multi-task setting of masked language modeling and categorial grammar supertagging. As an anonymous reviewer correctly pointed out, this method also satisfied the 3/4 requirements: (i), (ii), and (iv).
}
However, these approaches are potentially encoder-oriented and not suitable for unidirectional LMs; they assume the entire sentence as input and reflect the syntactic relationship between input words in a bottom-up manner.

Finally, another approach also closely aligned with the spirit of this research is a hybrid Parser and neural Language Model (PaLM; \citealp{peng2019PaLM}). PaLM is the integration of an unidirectional RNN LM with an additional attention layer, which would be supervised to attend the constituent spans among the spans ending at time $t-1$: $\{w_1,\cdots, w_{t-1}\},\cdots,\{w_{t-2}, w_{t-1}\}$. Although PaLM also meets the 3/4 requirements, it was by nature proposed for RNN LMs. The challenge arises when adapting PaLM to Transformer LMs; the generation of embeddings for the spans cannot be parallelized in a manner that is compatible with a self-attention mechanism of Transformer LMs.

To sum up, none of the previous approaches fully satisfy the requirements for our goal, highlighting the necessity for new methodologies.

\section{Proposed method: tree-planting}
\label{sec:TP}
\begin{figure*}
    \centering
    \includegraphics[width=15cm]{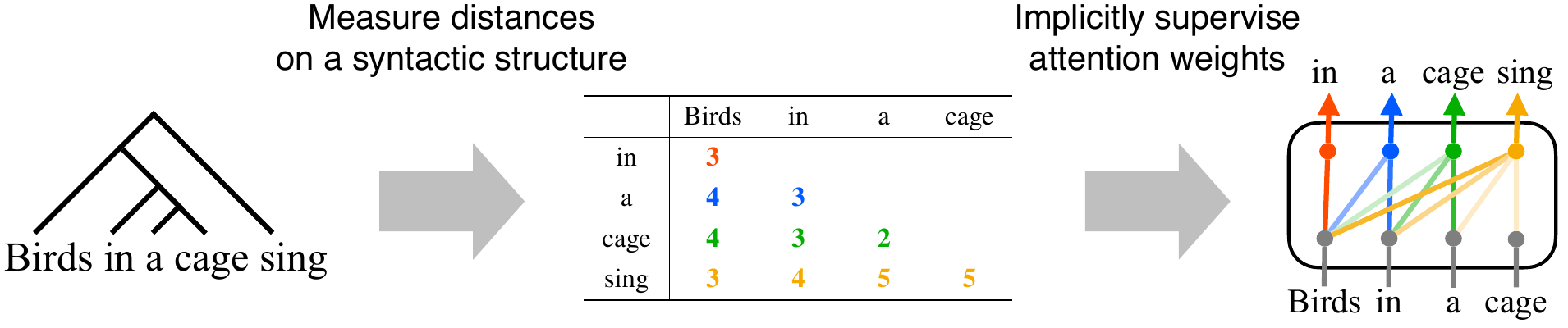}
    \caption{Overview of the proposed method: tree-planting}
    \label{fig:TP}
\end{figure*}
In this paper, we propose a new method dubbed \textbf{tree-planting}: we ``plant'' trees into attention weights of unidirectional Transformer LMs to reflect syntactic structures of natural language (Figure~\ref{fig:TP}). Specifically, unidirectional Transformer LMs trained with tree-planting will be called \textbf{Tree-Planted Transformers (TPT)}, which inherit the training efficiency from SLMs without changing the inference efficiency of their underlying Trans-
former LMs. Tree-planting is uniquely designed for unidirectional Transformer LMs, satisfying all the requirements discussed in \S\ref{subsec:const}: (i) parser-free inference, (ii) syntactic supervision, (iii) unidirectional LM, and (iv) parallel computation.

\subsection{Supervision of attention weights}
\label{subsec:supervision}
A self-attention mechanism of unidirectional Transformer LMs computes a representation for predicting the next token through a weighted sum of each token in the context. Specifically, when predicting the $i+1$-th token, the attention weights from the $i$-th token to the $j$-th token is computed as follows:
\begin{equation}
    \label{eq:attn}
    A_{ij} = \frac{\exp\left(\frac{\mathbf{Q}_i \mathbf{K}_j^T}{\sqrt{d_K}}\right)}{\sum_{k=1}^{i}\exp\left(\frac{\mathbf{Q}_i \mathbf{K}_k^T}{\sqrt{d_K}}\right)} ,
\end{equation}
where $\mathbf{Q}_i$ and $\mathbf{K}_j$ represent the query vector of the $i$-th token and the key vector of the $j$-th token, respectively, and $d_K$ denotes the dimension of the key vector. As Equation~\ref{eq:attn} shows, the computation for the $i+1$-th token prediction does not depend on any computation for the $1, \cdots, i$-th token predictions, which enables parallel computation.

In producing the supervision of attention weights, we extend the notion of \textit{syntactic distance array}~\citep{DBLP:conf/iclr/ShenLHC18,DBLP:conf/iclr/ShenTSC19,du2020Exploiting}, a 1D array of the number of edges on syntactic structures between two \textit{consecutive} words, to a 2D matrix between \textit{all pairs} of words:
\begin{equation}
    D_{ij} = \edge(w_i, w_j) ,
\end{equation}
where $w_i$ and $w_j$ represent the $i$-th and $j$-th words, respectively, and $\edge$ is the function that maps a pair of words to the number of edges on syntactic structures between them. This notion of \textit{syntactic distance matrix} could be applied to any kind of syntactic structure, as long as the number of edges can be counted on it.\footnote{Note importantly that compatibility with the syntactic distance matrix varies across types of syntactic structures because some kind of information would be lost during the conversion from syntactic structures to the matrix. For example, when applied to dependency structures, the information on the direction of syntactic dependency will be lost.}

Then, the syntactic distance matrix $\mathbf{D}$ is converted to the supervision of attention weights $\mathbf{S}$ as follows:
\begin{equation}
    S_{ij} = 
        \begin{cases}
            \frac{\exp(-D_{i+1,j})}{\sum_{k=1}^{i} \exp(-D_{i+1,k})} & (i \geq j)\\
            0 & (i < j)
        \end{cases} ,
\end{equation}
where $S_{ij}$ represents the supervision of the attention weight from the $i$-th word to the $j$-th word when predicting the $i+1$-th word. In short, this supervision expects the attention weight of each word to decrease exponentially with the number of edges between the predicted word.\footnote{We adopt an exponential function as \citet{lin2017Critical} reported that the mutual information between words would decay exponentially with respect to the number of edges on the syntactic structure between them.}

This supervision design is highly oriented towards unidirectional Transformer LMs; it considers syntactic structures involving the next word to be generated in a manner compatible with parallel computation of the self-attention mechanism. This only successfully satisfies the 3/4 requirements for our purpose: (ii) syntactic supervision, (iii) unidirectional LM, and (iv) parallel computation. To fulfill the remaining requirement of (i) parser-free inference, we adopt a strategy similar to that of \citet{strubell2018LinguisticallyInformed,chen2023Sudden}, designing the loss function to implicitly supervise attention.

\subsection{Loss function}
The supervision in \S\ref{subsec:supervision} is produced at the word level but Transformer LMs typically take their input at the subword level. To bridge this gap, we first convert the subword-level attention weight matrix $\mathbf{A}$ from a targeted Transformer LM to the word-level attention weight matrix $\mathbf{W}$ as follows:
\begin{align}
    &W_{ij} = \frac{C_{ij}}{\sum_{k=1}^{i} C_{ik}} ,\\
    &C_{ij} = \sum_{l=\rmstart(w_{i+1})}^{\rmend(w_{i+1})} \sum_{m=\rmstart(w_j)}^{\rmend(w_j)} A_{lm} ,
\end{align}
where $W_{ij}$ represents the word-level attention weight from the $i$-th word to the $j$-th word. $C_{ij}$ is defined as the sum of the subword-level attention weights over the subword inside $w_j$ when predicting the subword inside $w_{i+1}$, with $A_{lm}$ representing the subword-level attention weight from the $l$-th subword to the $m$-th subword and $\rmstart$ and $\rmend$ being the functions that map words to their start and end subword index, respectively. We employ $\mathbf{A}$ from specific attention heads called tree-planted heads.\footnote{\citet{qian2021Structural} also proposed the architecture which constrains some attention heads based on syntactic structures, or PLM-mask. PLM-mask and our tree-planting are similar in spirit, but they are quite different in their implementation: PLM-mask is a type of SLM that jointly generates a word sequence and its syntactic structure, but tree-planting builds TPTs, a type of LM. Furthermore, PLM-mask explicitly masks the attention weights based on the local parser state but tree-planting implicitly guides attention weights to reflect the whole syntactic structure.}

To implicitly supervise the word-level attention weight matrix $\mathbf{W}$ with the supervision $\mathbf{S}$, we introduce a tree-planting loss $\mathcal{L}_{\tree}$ employing a Kullback–Leibler (KL) Divergence loss $D_{\kl}$:\footnote{This loss function is inspired by \citet{ma2023DREEAM}, which guides attention weights to focus on relevant texts in a document-level relation extraction task.}

\begin{equation}
    \mathcal{L}_{\tree} = \frac{\sum_{i=1}^{n-1} D_{\kl}(\mathbf{S}_i||\mathbf{W}_i)}{n-1} ,
\end{equation}
where $n$ represents the length of a word sequence $\mathbf{w}$. In short, the tree-planting loss is the average KL Divergence loss in predicting each word except the beginning of $\mathbf{w}$.

During the training, $\mathcal{L}_{\tree}$ is averaged over tree-planted heads and balanced with the next word prediction loss $\mathcal{L}_{\mathrm{NWP}}$:
\begin{equation}
    \mathcal{L} = \mathcal{L}_{\nwp} + \lambda \frac{\sum_{h\in \mathcal{H}} \mathcal{L}_{\tree}^{(h)}}{H} ,
\end{equation}
where $\mathcal{L}_{\tree}^{(h)}$ represents a tree-planting loss for each tree-planted head $h$, $H$ is the total number of tree-planted heads, and $\lambda$ is a weight that balances the importance of the next word prediction loss and the average tree-planting loss. Unidirectional Transformer LMs trained with this loss function will be called Tree-Planted Transformers (TPT).

\section{Experiment}
\label{sec:exp}
To investigate whether TPTs can learn human-like syntactic knowledge as data-efficiently as SLMs while maintaining the modeling space of Transformer LMs unchanged, we conduct training on a small treebank and targeted syntactic evaluations on a syntactic knowledge benchmark.

\subsection{Settings}
\label{subsec:settings}
\paragraph{Training data}
We used \texttt{LG} dataset of \citet{hu2020Systematic}, which comprises approximately 1.8M sentences from BLLIP corpus~\citep{charniakeugene2000BLLIP}. Implicit syntactic supervision with each of three types of syntactic structures was investigated: (i) dependency structures (\texttt{[dep.]}), (ii) constituency structures (\texttt{[cons.]}), and (iii) binarized constituency structures (\texttt{[bin.]}). The (i) dependency structures were parsed with the \texttt{en\_core\_web\_sm} model from the \texttt{spacy} library~\citep{inesmontani2023Explosion}.\footnote{\url{https://spacy.io}} The (ii) constituency structures were re-parsed with the Berkeley Neural Parser~\citep{kitaev2018Constituency}\footnote{\url{https://github.com/nikitakit/self-attentive-parser}} by \citet{hu2020Systematic}. The (iii) binarized constituency structures were obtained by the binarization of the (ii) constituency structures with the \texttt{chomsky\_normal\_form} function from the \texttt{nltk} library~\citep{bird2009natural}.\footnote{\url{https://www.nltk.org}} We removed 43,986 sentences that the dependency parser analyzed as multiple sentences, but the constituency parser analyzed as a single sentence.

\paragraph{Models}
We used the same architecture and BPE tokenizer as GPT-2 small (124M; \citealp{radford2019Language}). The implementation of \texttt{GPT2LMHeadModel} and \texttt{GPT2Tokenizer} from the \texttt{transformers} library~\citep{wolf2020Transformers}\footnote{\url{https://huggingface.co/docs/transformers}} were employed but all parameters of \texttt{GPT2LMHeadModel} were randomly initialized. For the tree-planted head and the weight of the tree-planting loss, we adopted a single attention head on the last layer and $\lambda = 0.5$, respectively. The choice of the tree-planted head and the weight was based on preliminary experiments and the detailed effects of them will be described in \S\ref{sec:analysis}.

As baselines, we trained three models: (i) a model with zero weight for the tree-planting loss (\texttt{[zero]}), (ii) a model supervised with random syntactic distances that were generated from the distribution same as the dependency structures (\texttt{[rand.]}), and (iii) a model supervised with sequential distances (\texttt{[seq.]}). Note importantly, (i) is equivalent to a Transformer LM. Hyperparameters are shown in Appendix~\ref{app:params}.

\paragraph{Evaluation data}
We evaluated syntactic knowledge of the models via targeted syntactic evaluations on the SyntaxGym benchmark~\citep{gauthier2020SyntaxGym}. The SyntaxGym benchmark comprises six syntactic \textit{circuits}: \texttt{Agreement}, \texttt{Center Embedding}, \texttt{Garden-Path Effects}, \texttt{Gross Syntactic States}, \texttt{Licensing}, and \texttt{Long-Distance Dependencies}. Each syntactic circuit consists of 2--10 syntactic \textit{suites} on a specific type of syntactic phenomenon; for example, the \texttt{Agreement} circuit contains syntactic suites such as ``subject-verb number agreement with a prepositional phrase''. Each syntactic suite contains 20--30 syntactic \textit{items} with different vocabulary; for example, the ``subject-verb number agreement with a prepositional phrase'' suite includes syntactic items as follows:
\eenumsentence[1]{
    \item \ac The author next to the senators \underline{is} good.
    \item \un The author next to the senators \underline{are} good.
}
LMs' predictions are evaluated against \textit{success criterion}, which specifies the inequality between conditions within an item; for example, the underlined position of the grammatical sentence (\ex{1}a) should be assigned the higher conditional probability than the ungrammatical one (\ex{1}b). 

All models were trained and evaluated two times with different random seeds. We report average accuracies with a standard deviation, along with word-level perplexity on the BLLIP test set.

\subsection{Overall accuracies}
\begin{table}[t]
    \centering
    \begin{tabular}{lcc}
        \hline
         & SG ($\uparrow$) & PPL ($\downarrow$)\\
         \hline
         \hline
         \multicolumn{3}{l}{\textbf{Baselines:}} \\
         TPT\texttt{[zero]} & 71.7 $\pm$ 0.3 & 47.5 $\pm$ 0.1$\spadesuit$ \\
         TPT\texttt{[rand.]} & 69.0 $\pm$ 1.0 & 47.4 $\pm$ 0.1$\spadesuit$ \\
         TPT\texttt{[seq.]}  & 70.1 $\pm$ 3.5 & \textbf{47.3 $\pm$ 0.2}$\spadesuit$ \\
         \hline
         \multicolumn{3}{l}{\textbf{TPTs (ours):}} \\
         TPT\texttt{[dep.]} & \textbf{77.1 $\pm$ 0.2} & 47.7 $\pm$ 0.1$\spadesuit$ \\
         TPT\texttt{[cons.]} & 75.8 $\pm$ 0.0 & \textbf{45.5 $\pm$ 0.0}$\heartsuit$ \\
         TPT\texttt{[bin.]} & 73.0 $\pm$ 1.8 & 45.6 $\pm$ 0.2$\heartsuit$ \\
         \hline
         \multicolumn{3}{l}{\textbf{SLMs (comparable):}} \\
         PLM & 42.2 $\pm$ 1.2 & - \\
         PLM-mask & 42.5 $\pm$ 1.5 & - \\
         \hline
         \hline
         \multicolumn{3}{l}{\textbf{SLMs (reference):}} \\
         PLM\dag & 73.2 $\pm$ 0.6 & 49.3 $\pm$ 0.3$\heartsuit$ \\
         PLM-mask\dag & 74.6 $\pm$ 1.0 & 49.1 $\pm$ 0.3$\heartsuit$ \\
         TG\ddag & 82.5 $\pm$ 1.6 & 30.3 $\pm$ 0.5$\heartsuit$ \\
         \hline
         \multicolumn{3}{l}{\textbf{LLM-like models (reference):}} \\
         GPT-2\P & 78.4 & - \\
         Gopher\P & 79.5 & - \\
         Chinchilla\P & 79.7 & - \\
         \hline
    \end{tabular}
    \caption{Overall accuracies of TPTs and their baselines on the SyntaxGym benchmark (SG), along with word-level perplexity on the BLLIP test set (PPL). The overall accuracies were calculated across the syntactic suites. $\dag$ and $\ddag$ represent the reference points as their inference methods are more costly than TPTs. $\P$ are also the reference points as they were trained on significantly larger corpora than TPTs. Perplexity can be directly comparable only within the same mark, either $\spadesuit$ or $\heartsuit$, due to differences in the tokenization of the constituency parser and dependency parser.}
    \label{tab:main_result}
\end{table}
Table~\ref{tab:main_result} shows the overall accuracies of TPTs and their baselines on the SyntaxGym benchmark (SG), along with word-level perplexity on the BLLIP test set (PPL). The overall accuracies were calculated across the syntactic suites.
We also report the accuracies of several SLMs that were also trained on the same BLLIP-\texttt{LG} dataset: PLM, PLM-mask~\citep{qian2021Structural}, and TG~\citep{sartran2022Transformer}. Only unmarked PLM and PLM-mask can be fairly comparable with TPTs as their evaluation was conducted generating a single syntactic structure via greedy search, to align inference costs with TPTs.\footnote{The fair comparison of TG was not performed because their trained parameters were not publicly available.} $\dag$ and $\ddag$ represent the reference points from \citet{sartran2022Transformer} as their inference methods are more costly than TPTs: $\dag$ and $\ddag$ employed word-synchronous beam search~\citep{stern2017Effective} of action beam size 100\footnote{Word beam size was 10 and fast track size was 5.} and the external parser~\citep{dyer2016Recurrent} to generate 300 candidate structures, respectively. The accuracies of several LLM-like models are also reported from \citet{sartran2022Transformer}: GPT-2~\citep{radford2019Language}, Gopher~\citep{rae2022Scaling}, and Chinchilla~\citep{hoffmann2022Training}. They are also the reference points as these LLM-like models were trained on $250\times$ to $1000\times$ larger corpora (denoted by $\P$). Perplexity can be directly comparable only within the same mark, either $\spadesuit$ or $\heartsuit$, due to differences in the tokenization of the dependency parser and constituency parser.

There are some important observations in the overall accuracies on the SyntaxGym benchmark:
\begin{itemize}
    \item TPT\texttt{[zero]}, which is equivalent to a Transformer LM, underperformed all TPTs with some implicit syntactic supervision, suggesting that tree-planting can induce data-efficient syntactic generalization.
    \item TPTs\texttt{[rand.][seq.]} also underperformed all TPTs with some implicit syntactic supervision, indicating that not KL Divergence loss itself but the loss based on \textit{syntactic structures} is necessary.
    \item Among TPTs with some implicit syntactic supervision, TPT\texttt{[dep.]} achieved the best performance. We further investigate this point in \S\ref{subsec:circuit}.
    \item Most importantly, despite the lack of explicit generation of syntactic structures, TPTs\texttt{[dep.][cons.]} significantly outperformed not only the comparable SLMs (unmarked PLM and PLM-mask) but also the various SLMs that generate hundreds of syntactic structures in parallel (PLM$\dag$ and PLM-mask$\dag$).
\end{itemize}
Even though the best TPT\texttt{[dep.]} underperformed the reference points of TG, which consumed at least $300\times$ higher inference cost, and LLM-like models, which were trained on at least $250\times$ larger data, these observations adequately suggest that TPTs can learn human-like syntactic knowledge as data-efficiently as SLMs while maintaining the modeling space of Transformer LMs unchanged.

Regarding perplexity, although TPT\texttt{[dep.]} numerically underperformed its comparable baselines, they all achieved similar perplexity with no significant differences.

\subsection{Circuit accuracies}
\label{subsec:circuit}
\begin{figure}
    \centering
    \includegraphics[width=7.5cm]{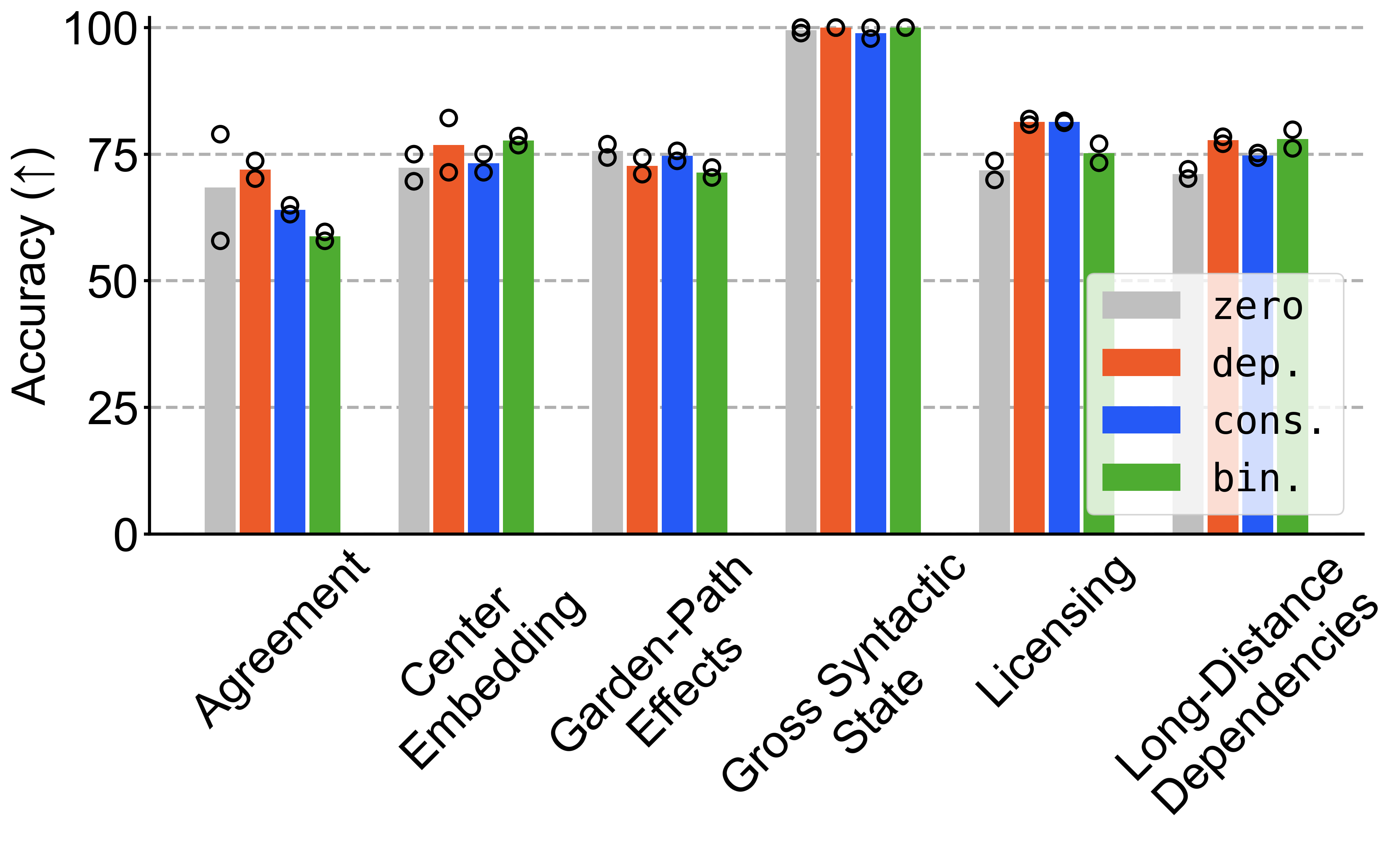}
    \caption{Circuit accuracies of TPTs with some implicit syntactic supervision and the baseline model with zero weight for the tree-planting loss on the SyntaxGym benchmark. The circuit accuracies calculated across the syntactic suites (the vertical axis) are plotted against the models (the horizontal axis), with each dot representing the accuracy of a specific seed.}
    \label{fig:SG}
\end{figure}
In this subsection, we investigate the reason for the high compatibility between tree-planting and \textbf{dependency structures} through the lens of circuit accuracies.
Figure~\ref{fig:SG} shows the circuit accuracies of TPTs with some implicit syntactic supervision and the baseline model with zero weight for the tree-planting loss on the SyntaxGym benchmark. The circuit accuracies calculated across the syntactic suites (the vertical axis) are plotted against the models (the horizontal axis), with each dot denoting the accuracy of a specific seed.

\paragraph{vs. zero supervision}
TPT\texttt{[dep.]} outperformed TPT\texttt{[zero]} on 5/6 circuits, suggesting that tree-planting with dependency structures is generally advantageous over zero supervision. However, the \texttt{Garden-Path Effects} circuit presents an exception, where LMs are evaluated for the ability to be surprised in a human-like manner, through comparisons between sentences minimally different not in \textit{grammaticality} but in \textit{local ambiguity}~\citep{hu2020Systematic}. The underperformance of TPT\texttt{[dep.]} may suggest that due to the syntactic knowledge introduced by tree-planting with dependency structures, TPT\texttt{[dep.]} was no longer surprised by locally ambiguous but grammatical sentences. We further investigate this point in Appendix~\ref{app:disadvantage}.

\paragraph{vs. constituency structures}
Surprisingly, on 5/6 circuits, TPT\texttt{[dep.]} outperformed TPT\texttt{[cons.]}. The only exception is the \texttt{Garden-Path Effects} circuit, where the potential disadvantage of tree-planting with dependency structures exists, as mentioned above. Specifically, TPT\texttt{[dep.]} most significantly outperformed TPT\texttt{[cons.]} on the \texttt{Agreement} circuit, which includes the syntactic items such as (\ex{1}) from \S\ref{subsec:settings}: ``The author next to the senator \underline{is/*are} good''. For these syntactic items, only the head of the subject NP (\textit{author}) is always nearest to the main verb (\textit{is}/*\textit{are}) on dependency structures, but the same does not hold on constituency structures: in constituency structures, the determiner of the subject NP (\textit{the}) and the head of the post-modifying PP (\textit{to}) are as nearest to the main verb as the head of the subject NP (cf. Appendix~\ref{app:example}). Given that tree-planting utilizes the number of edges as implicit syntactic supervision, the property of dependency structures may potentially be more desirable for tree-planting than constituency structures.

\paragraph{vs. binarized constituency structures}
TPT\texttt{[dep.]} outperformed TPT\texttt{[bin.]} on 3/6 circuits, with similar performance (a difference less than $-1.0\%$) on the other 3 circuits. Notably, TPT\texttt{[dep.]} achieved significantly better performance (a difference more than $+5.0\%$) on the \texttt{Agreement} and \texttt{Licensing} circuits. 
\citet{noji2023How} reported that deep syntactic supervision is not always optimal; rather mild syntactic supervision is sufficient for addressing long-distance dependencies between elements within and outside complex NP subjects. Given that (i) the \texttt{Agreement} and \texttt{Licensing} circuits consist only of syntactic suites that exemplify this condition\footnote{Among the other syntactic circuits, the \texttt{Center Embedding} circuit also consist only of syntactic suites that exemplify this condition.} and (ii) the average syntactic distance in the training data is significantly shorter for dependency structures (4.8) than binarized constituency structures (13.1), it could be argued that tree-planting would be more ``good enough'' syntactic supervision with dependency structures, rather than with binarized constituency structures.\footnote{The average syntactic distance of constituency structures is 10.0. This suggests that tree-planting would also be more ``good enough'' syntactic supervision with dependency structures rather than with constituency structures, besides the points discussed in the ``vs. constituency structures'' paragraph.}

\section{Analysis}
\label{sec:analysis}
In this section, we report the effects of (i) the number of tree-planted heads and (ii) the weight of a tree-planting loss, using TPT\texttt{[dep.]}.
\begin{figure*}
    \centering
    \includegraphics[width=15cm]{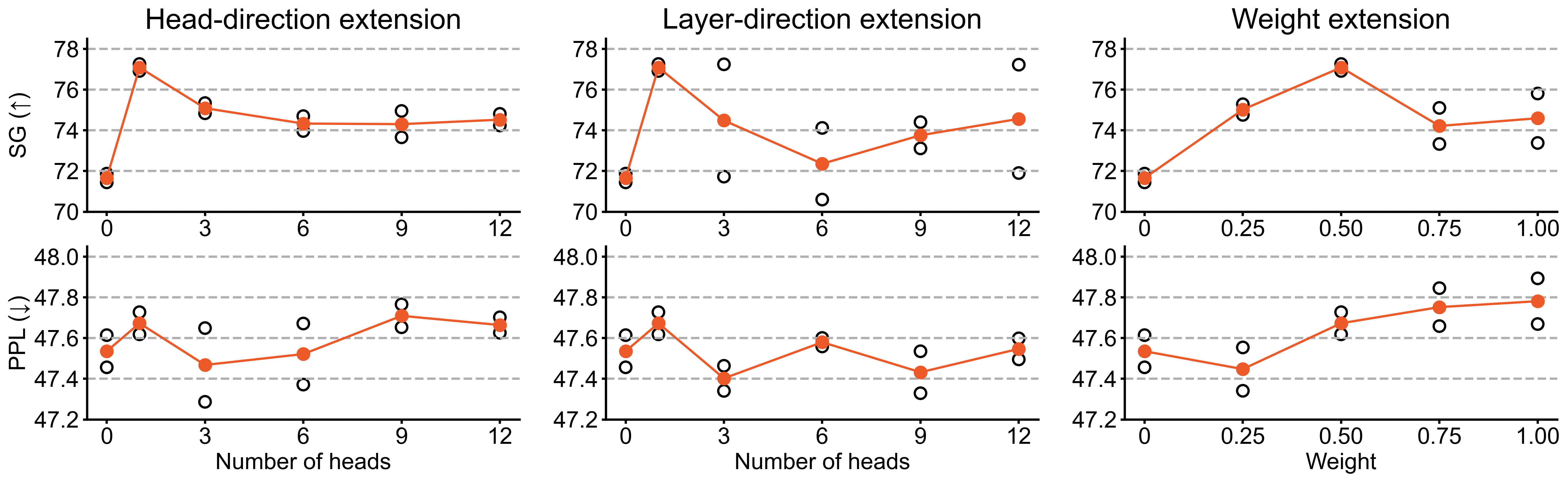}
    \caption{The results of the head-direction, layer-direction, and weight extension. For the head-direction and layer-direction extension, the overall accuracies on the SyntaxGym benchmark and the perplexity on the BLLIP test set (the vertical axis) are plotted against the number of tree-planted heads (the horizontal axis). For the weight extension, the horizontal axis indicates the weight of the tree-planting loss.
    }
    \label{fig:ana_hlw}
\end{figure*}
\subsection{Number of tree-planted heads}
\label{subsec:number}
Our TPTs are based on a 12-layer, 12-head Transformer LM. In \S\ref{sec:exp}, out of $12\times12$ heads, we adopted a single attention head on the last layer as a tree-planted head. In this subsection, we explore two alternatives: (i) head-direction extension and (ii) layer-direction extension. For the head-direction extension, $H=0, 1, 3, 6, 9, 12$ heads on the last layer were adopted as tree-planted heads. For the layer-direction extension, one attention head from each of the bottom $H=0, 1, 3, 6, 9, 12$ layers was adopted as tree-planted heads.

In the left two columns of Figure~\ref{fig:ana_hlw}, the results of the head-direction and layer-direction extension are shown: the overall accuracies on the SyntaxGym benchmark (SG) and the word-level perplexity on the BLLIP test set (PPL) (the vertical axis) are plotted against the number of tree-planted heads (the horizontal axis). Each dot denotes the accuracy or perplexity of a specific seed. For both settings, $H=0,1$ are equivalent to TPT\texttt{[zero]} and TPT\texttt{[dep.]}, respectively.

Considering the overall accuracies on the SyntaxGym benchmark, in both the head-direction and layer-direction extension, the highest accuracy was achieved when only a single head was adopted as a tree-planted head, while it is noteworthy that all the models with tree-planted heads outperformed the model without them. Incidentally, it should be mentioned that the result of the layer-direction extension exhibited significantly more variability. Although the exact reason why a single tree-planted head would work well is unclear, the adoption of multi tree-planted heads inherently induces the handling of redundant information across heads, which might potentially hinder the management of non-syntactic information of natural languages (e.g., lexical information). Regarding perplexity, no consistent trend emerged.

\subsection{Weight of a tree-planting loss}
\label{subsec:weight}
In \S\ref{sec:exp}, we adopted $\lambda=0.5$ as the weight of the tree-planting loss. Here, we extend $\lambda$ to $0.0$, $0.25$, $0.50$, $0.75$, and $1.00$. $\lambda=0,0.50$ are equivalent to TPT\texttt{[zero]} and TPT\texttt{[dep.]}, respectively.

The rightmost column of Figure~\ref{fig:ana_hlw} shows the results of the weight extension. The overall accuracies on the SyntaxGym benchmark display a single-peaked pattern, with the maximum reached for $\lambda=0.50$. Interestingly, this result suggests that by overtly focusing on reflecting syntactic structures, TPTs paradoxically become unable to learn syntactic knowledge efficiently. On the other hand, we observed that the perplexity got worse monotonically as the weight increased. From these observations, we may deduce that to acquire syntactic knowledge efficiently, TPTs should learn not only to reflect syntactic structures in their attention weights but also to precisely predict the next word. Therefore, the weight of the tree-planting loss emerges as a critical hyperparameter, indicating that the search for the optimal balance between the next-word prediction loss and tree-planting loss is vital for developing more human-like TPTs.


\section{Conclusion}
In this paper, we propose a new method dubbed \textbf{tree-planting}: instead of explicitly generating syntactic structures, we ``plant'' trees into attention weights of unidirectional Transformer LMs to implicitly reflect syntactic structures of natural language. Specifically, unidirectional Transformer LMs trained with tree-planting will be called \textbf{Tree-Planted Transformers (TPT)}, which inherit the training efficiency from SLMs without changing the inference efficiency of their underlying Transformer LMs. Targeted syntactic evaluations on the SyntaxGym benchmark demonstrated that TPTs, despite the lack of explicit generation of syntactic structures, significantly outperformed not only vanilla Transformer LMs but also various SLMs that generate hundreds of syntactic structures in parallel. This result suggests that TPTs can learn human-like syntactic knowledge as data-efficiently as SLMs while maintaining the modeling space of Transformer LMs unchanged.

\section*{Limitations}
This paper has at least three limitations. First, we only conducted sentence-level tree-planting. Typically, Transformer LMs are trained at the document level, but SLMs are trained at the sentence level~\citep{dyer2016Recurrent,kuncoro2017What,noji2021Effective,yoshida2022Composition}, because on treebanks the annotations are assigned at the sentence level. Because of this constraint, we also employed sentence-level experimental design and verified the effectiveness of the proposed method first and foremost. Recent research in SLMs, however, has begun to extend treebank annotations to the document level and train document-level SLMs on them~\citep{sartran2022Transformer,murty2023Pushdown}. When constructing TPTs for practical use, it might be beneficial to follow these recent studies and perform tree-planting with document-level annotations.

Second, we only evaluated TPTs on the syntactic knowledge benchmark and perplexity. Recently, \citet{murty2023Pushdown} evaluated the performance of SLMs on text classification tasks; to the best of our knowledge, this is the first work that evaluated SLMs on tasks other than the targeted syntactic evaluations. More recently, \citet{hu2024Generative} evaluated unsupervised SLMs on text generation tasks as well as text classification tasks. \citet{murty2023Pushdown} and \citet{hu2024Generative} both suggested that syntactic supervision could also be beneficial to solving them; this indicates that there is also room for a broader evaluation of our methodology.

Finally, there might still be room for further improvement of tree-planting and TPTs. For instance, an in-depth comparison between TPT and TG, which achieved much better accuracy on the \texttt{Center Embedding} and \texttt{Garden-Path Effects} circuits \citep[cf.][]{sartran2022Transformer}, could provide insights for improving the design of tree-planting. Additionally, since the modeling space of TPT is identical to that of underlying Transformer LMs, TPTs are theoretically capable of continual learning on standard text corpora. In future work, we plan to develop a novel method to scale TPTs on large text corpora without compromising syntactic knowledge.


\section*{Ethical considerations}
A significant feature of TPT lies in the training efficiency and inference efficiency, which can potentially contribute to reducing computational resources. One minor concern is the possibility of bias in the models utilized in this paper, attributed to the training data (i.e., the BLLIP corpus), although this experimental setting follows conventional practices in the literature on SLMs. We employed ChatGPT and Grammarly for writing assistance and utilized ChatGPT and Copilot for the development of experimental code. These tools were used in compliance with the ACL 2023 Policy on the Use of AI Writing Assistance.

\section*{Acknowledgements}
We appreciate the insightful reviews provided by the three anonymous ARR reviewers. We would also like to thank Peng Qian for supplying the re-parsed BLLIP-\texttt{LG} dataset, which was used to train TPT\texttt{[cons.]}. Special thanks to Laurent Sartran for answering various questions regarding \citet{sartran2022Transformer}. We are grateful to Kohei Kajikawa, Shinnosuke Isono, Yushi Sugimito, and Taketeru Yamakoshi for their valuable comments and suggestions. This work was supported by JSPS KAKENHI Grant Number 24H00087, Grant-in-Aid for JSPS Fellows JP24KJ0800, JST PRESTO Grant Number JPMJPR21C2, and JST SPRING Grant Number JPMJSP2108.

\bibliography{Lab,anthology}
\clearpage
\appendix
\section{Hyperparameters}
\label{app:params}
Hyperparameters for our experiments are shown in Table~\ref{tab:params}, which primarily followed default settings. All models were trained and evaluated on $8\times$ NVIDIA V100 (16GB). The total computational cost for all experiments in this paper amounted to about 1,300 GPU hours.

\begin{table}
    \centering
    \begin{tabular}{lc}
        \hline
         Optimizer & AdamW \\
         Learning rate & 5e-5 \\
         Number of epochs & 10 \\
         Dropout rate & 0.1 \\
         Batch size & 256 \\
         \hline
    \end{tabular}
    \caption{Hyperparameters for our experiments}
    \label{tab:params}
\end{table}

\section{Further investigation of the \texttt{Garden-Path Effects} circuit}
\label{app:disadvantage}
In \S\ref{subsec:circuit}, we suggest the probability that syntactic knowledge introduced by tree-planting with dependency structures may prevent TPT\texttt{[dep.]} from being surprised by locally ambiguous but grammatical sentences. To inspect this, we break down the \texttt{Garden-Path Effects} circuit into the syntactic suites: ``main verb / reduced relative clause'' (MVRR) and ``NP/Z garden-paths'' (NP/Z).

\begin{figure}
    \centering
    \includegraphics[width=7.5cm]{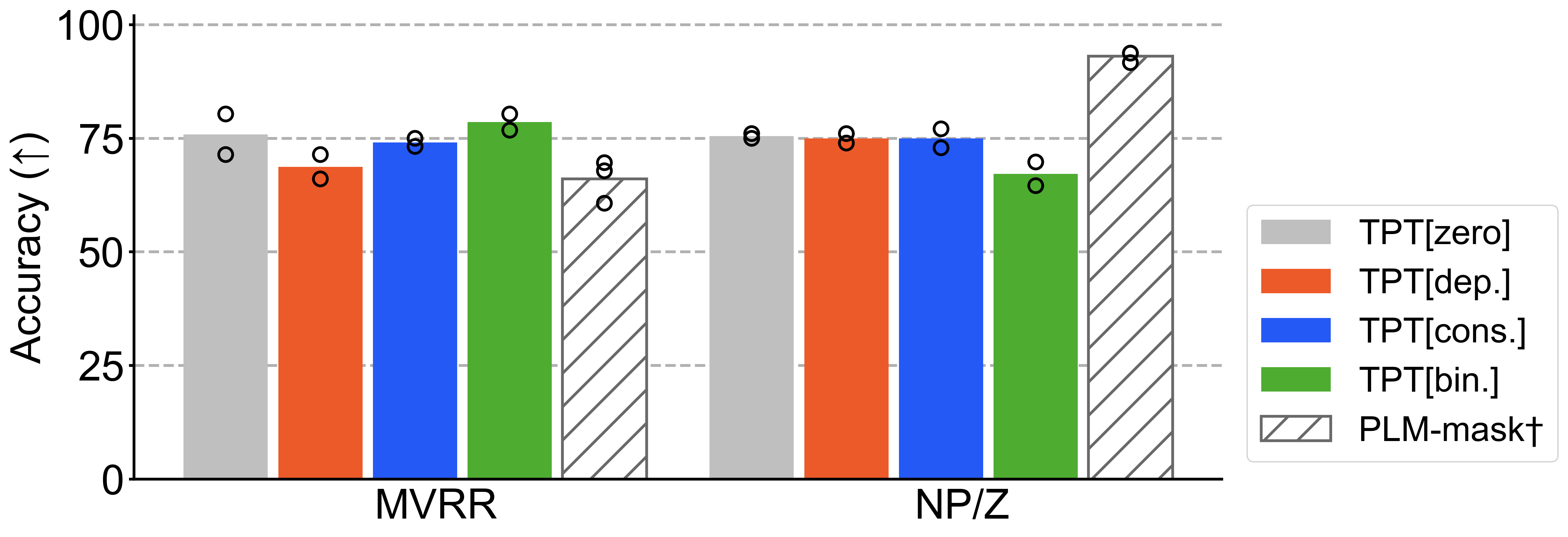}
    \caption{Suite accuracies of TPTs with some implicit syntactic supervision and the baseline model with zero weight for the tree-planting loss on the \texttt{Garden-Path Effects} circuit, with the reference point of the more inference-costly SLM, or PLM-mask$\dag$~\citep{qian2021Structural}}
    \label{fig:sg_gpe}
\end{figure}

Figure~\ref{fig:sg_gpe} shows the suite accuracies of TPTs with some implicit syntactic supervision and the baseline model with zero weight for the tree-planting loss on the \texttt{Garden-Path Effects} circuit, with the reference point of the more inference-costly SLM, or PLM-mask$\dag$~\citep{qian2021Structural}. We find that the deficiency of TPT\texttt{[dep.]} is attributed to its inadequate performance on the MVRR circuit, which includes the syntactic items as follows:
\eenumsentence[2]{
    \item \acexc The dog seen on the beach \underline{chased} \\\acexc after a bird.
    \item \exc The dog walked on the beach \underline{chased} \\\acexc after a bird.
}
The success criterion on these suites defines that the underlined position of the unambiguous sentence (\ex{2}a) should be assigned a higher conditional probability than the locally ambiguous one (\ex{2}b). We speculate that TPT\texttt{[dep.]} might lose its sensitivity to the local ambiguity introduced by the participle verb (\textit{seen/walked}), as it is guided to focus more intently on the head of the subject NP (\textit{dog}) when predicting the main verb (\textit{chased}), than the unrestricted baseline.

Conversely, TPT\texttt{[cons.][bin.]} did not underperform TPT\texttt{[zero.]} on the MVRR suites. This result could be straightforwardly understood, given that on these structures, the participle verb (\textit{seen/walked}) and the head of the subject NP (\textit{dog}) are equidistant from the main verb (\textit{chased}). However, it is worth noting that the determiner of the subject NP (\textit{the}) also shares this distance, which may not always be a desirable property for tree-planting (cf. \S\ref{subsec:circuit}).

Finally, PLM-mask$\dag$, the more inference-costly SLM, also underperformed TPT\texttt{[zero]} on the MVRR suites. This suggests that the models with explicit syntactic supervision may also struggle with losing sensitivity to the local ambiguity as PLM\texttt{[dep.]}.

\clearpage
\section{Dependency/constituency structures of \texorpdfstring{(\ex{1})}{(1)} from \S\ref{subsec:settings}}
\label{app:example}
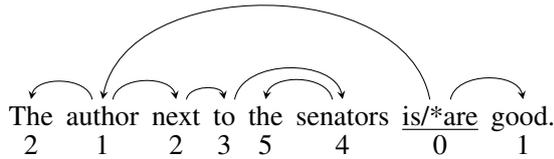
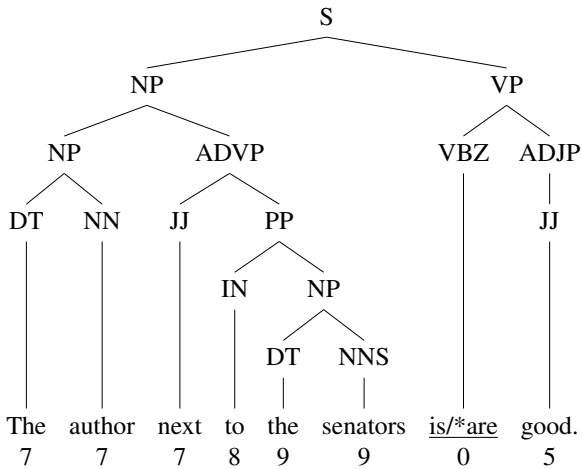
\begin{figure}[H]
\centering
\begin{subfigure}{\columnwidth}
\centering
\resizebox{\columnwidth}{!}{%
    \begin{dependency}[theme = simple]
    \begin{deptext}[column sep=0em]
    The \& author \& next \& to \& the \& senators \& \underline{is/*are} \& good.\\
    2 \& 1 \& 2 \& 3 \& 5 \& 4 \& 0 \& 1\\
    \end{deptext}
    \depedge{2}{1}{}
    \depedge{7}{2}{}
    \depedge{2}{3}{}
    \depedge{3}{4}{}
    \depedge{6}{5}{}
    \depedge{4}{6}{}
    \depedge{7}{8}{}
    \end{dependency}
}
\caption{Dependency structure}
\label{subfig:dependency}
\end{subfigure}

\vspace{1cm} 

\begin{subfigure}{\columnwidth}
\centering
\resizebox{\columnwidth}{!}{%
    \begin{forest}
    for tree={
        align=center, 
        parent anchor=south, 
        child anchor=north,
        l=0.1cm, 
        s=0.1cm, 
        inner sep=1pt, 
    },
    [S, for tree={parent anchor=south, child anchor=north}
      [NP
        [NP [DT [The\\7, tier=word]] [NN [author\\7, tier=word]]]
        [ADVP [JJ [next\\7, tier=word]] [PP [IN [to\\8, tier=word]] 
          [NP [DT [the\\9, tier=word]] [NNS [senators\\9, tier=word]]]
        ]]
      ]
      [VP [VBZ [\underline{is/*are}\\0, tier=word]] [ADJP [JJ [good.\\5, tier=word]]]]
    ]
    \end{forest}
}
\caption{Constituency structure}
\label{subfig:constituency}
\end{subfigure}

\caption{Dependency/constituency structures of (\ex{1}) from \S\ref{subsec:settings}}
\label{fig:structures}

\end{figure}

To assist the discussion in \S\ref{subsec:circuit}, the dependency and constituency structures of (\ex{1}) from \S\ref{subsec:settings} were displayed in Figure~\ref{subfig:dependency} and \ref{subfig:constituency}, respectively. Numbers below each word represent the number of edges from the underlined position. To parse (\ex{1}), the parsers referenced in \S\ref{subsec:settings} were employed.


\section{Begin/End Of Sentence Tokens}
Sentences in the BLLIP corpus do not include Begin/End of Sentence (\texttt{BOS}/\texttt{EOS}) tokens, which are essential for sequences processed by LMs. To integrate these tokens, we implemented the following modifications:
\begin{itemize}
    \item For dependency structures, we introduced \texttt{BOS}/\texttt{EOS} tokens by defining new edges from the \texttt{ROOT} to these tokens.
    \item For constituency structures, we introduced \texttt{BOS}/\texttt{EOS} tokens by modifying the tree structure to encapsulate the original structure within a new root node, specifically by adding a \texttt{BOS} token and an \texttt{EOS} token as the first and the last child of this new root, respectively.
\end{itemize}

\section{License of the data/tools}
We summarize the license of the data/tools employed in this paper in Table~\ref{tab:license}. All data and tools were used under their respective license terms.
\begin{table}[H]
    \small
    \centering
    \begin{tabular}{p{5cm}p{2cm}}
    \hline
    Data/tool     &  License\\
    \hline
    BLLIP~\citep{charniakeugene2000BLLIP}&BLLIP 1987-89 WSJ Corpus Release 1 License Agreement\\
    SyntaxGym~\citep{gauthier2020SyntaxGym}& MIT\\
    \hline
    \texttt{spacy}~\citep{inesmontani2023Explosion}& MIT\\
    \texttt{nltk}~\citep{bird2009natural} & Apache 2.0\\
    \texttt{transformers}~\citep{wolf2020Transformers} & Apache 2.0\\
    \hline
    Berkeley Neural Parser~\citep{kitaev2018Constituency} & MIT\\
    PLM/PLM-mask~\citep{qian2021Structural} & MIT\\
    \hline
    \end{tabular}
    \caption{License of the data/tools}
    \label{tab:license}
\end{table}

\end{document}